\pdfoutput=1

\documentclass[11pt]{article}

\usepackage[preprint]{acl}

\usepackage{times}
\usepackage{latexsym}

\usepackage[T1]{fontenc}

\usepackage[utf8]{inputenc}

\usepackage{microtype}

\usepackage{inconsolata}

\usepackage{amsmath}

\usepackage{graphicx}

\usepackage{hyperref}
\usepackage{listings} 
\usepackage{xcolor} 

\usepackage{arydshln}

\usepackage{booktabs}

\usepackage{multirow}

\definecolor{codegreen}{rgb}{0,0.6,0}
\definecolor{codegray}{rgb}{0.5,0.5,0.5}
\definecolor{codepurple}{rgb}{0.58,0,0.82}
\definecolor{backcolour}{rgb}{0.95,0.95,0.92}
\definecolor{codemarine}{rgb}{0.0,0.0,0.5}

\lstdefinestyle{mystyle}{
    backgroundcolor=\color{backcolour},   
    commentstyle=\color{codegreen},
    keywordstyle=\color{magenta},
    numberstyle=\tiny\color{codegray},
    stringstyle=\color{codepurple},
    basicstyle=\footnotesize\ttfamily,
    breakatwhitespace=false,         
    breaklines=true,                 
    captionpos=b,                    
    keepspaces=true,                 
    numbers=left,                    
    numbersep=5pt,                  
    showspaces=false,                
    showstringspaces=false,
    showtabs=false,                  
    tabsize=2,
    frame=single,                    
    morekeywords={self},             
    keywordstyle=\color{codemarine}, 
}

\title{Enhancing LLM Factual Accuracy with RAG to Counter Hallucinations: A Case Study on Domain-Specific Queries in Private Knowledge-Bases}

\author{
    Jiarui Li\thanks{These authors contributed equally to this work.}, 
    Ye Yuan\footnotemark[1], 
    Zehua Zhang\footnotemark[1] \\
    Information Network Institute, \\
    Carnegie Mellon University \\
    \texttt{\{jiaruil3, yeyuan3, zehuazha\}@andrew.cmu.edu} \\
}

\begin{document}
\maketitle 

\begin{abstract}
\vspace{-0.5em}
We proposed an end-to-end system design towards utilizing Retrieval Augmented Generation (RAG) to improve the factual accuracy of Large Language Models (LLMs) for domain-specific and time-sensitive queries related to private knowledge-bases. Our system integrates RAG pipeline with upstream datasets processing and downstream performance evaluation. Addressing the challenge of LLM hallucinations, we finetune models with a curated dataset which originates from CMU's extensive resources and annotated with the teacher model. Our experiments demonstrate the system's effectiveness in generating more accurate answers to domain-specific and time-sensitive inquiries. The results also revealed the limitations of fine-tuning LLMs with small-scale and skewed datasets. This research highlights the potential of RAG systems in augmenting LLMs with external datasets for improved performance in knowledge-intensive tasks. Our code and models are available on \href{https://github.com/anlp-team/LTI_Neural_Navigator}{Github}.
\vspace{-0.5em}
\end{abstract}

\section{Introduction}

Large Language Models (LLMs) such as \texttt{ChatGPT} and \texttt{LLaMA-2} have demonstrated their efficacy in question-answering (QA) tasks\cite{GPT3, LLAMA2}. These models are trained on vast amounts of text data from various sources, enabling them to understand and generate human-like responses to questions across a wide range of topics. While LLMs acquired such ability through extensive pretraining, it is evident that information of some scale is retained in LLMs' parameters\cite{StarCoder}. LLMs are known to hallucinate: the model may generate results that are irrelevant or random outputs that may hold no correlation with inputs or desired outputs \cite{Hallucination_survey}. Hallucination is often unreliable and may be misleading or offensive to the users. Many reasons are proposed to explain LLM hallucination \cite{zhu2024halueval, bender2021dangers, burns2022discovering}. While pretraining should cultivate the model's emergent ability, it is not expected to help the model remember all the pretraining data accurately. Recently, methods such as In-Context Learning (ICL) have been investigated to discover LLM's performance on certain tasks such as sentiment classification in a few-shots setting \cite{dong2022survey, wei2023larger, yousefi2023decoding}. By providing a few examples related to the desired tasks, the model could solve the expected problem with relatively higher accuracy without the need to change its parameters. Retrieval Augmented Generation (RAG) \cite{RAG} is recently regarded as a promising solution to extend the model's generation capacity on tasks that require high factuality by introducing context information from external datasets. Connecting the models to external datasets, RAG could help LLMs handle knowledge-intensive tasks and allow continuous knowledge updates and integration of domain-specific information. 


Our project aims to build a system which outperforms baseline open-source solutions. In detail, our contributions are as follows.

\begin{enumerate}
\vspace{-0.5em}
    \item \textbf{Creation of a Specialized Dataset}: We curated a comprehensive and accurate dataset focused on Carnegie Mellon University (CMU) and the Language Technology Institute (LTI), using a novel Web crawler and automated annotation process.
    \vspace{-0.5em}
    \item \textbf{Development of a RAG Pipeline}: We innovated a RAG pipeline with SOTA research findings, which ensures advanced linguistic pattern handling capabilities.
    \vspace{-0.5em}
    \item \textbf{Extensive Experiment-based Evaluation}: Through rigorous experiments, including ablation studies and case studies, we demonstrated the performance of our system. 
\end{enumerate}

\begin{figure*}[t]
\centering
\includegraphics[width=\textwidth]{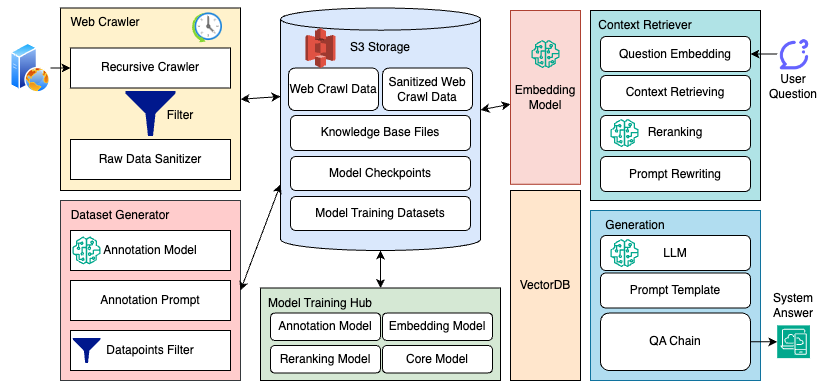}
\caption{Overview of our system design. The process begins with a \textbf{Web Crawler}, which consists of a Recursive Crawler and a Filter to gather raw data. This raw data is sanitized and stored in an S3 storage bucket. The \textbf{Dataset Generator} relies on an Annotation Model to prompt annotations and filter datapoints, generating finetune datasets for all models in the system, i.e. Annotation, Embedding, Reranking, and Core Models. For query processing, the \textbf{Context Retriever} embeds the user question and retrieves relevant contexts, which are re-ranked and rewritten if necessary. The \textbf{Generation module} then utilizes the Core Model and a Prompt Template to generate a system answer through a QA Chain, resulting in an answer that leverages both retrieved information and generative capabilities for accurate and context-aware responses.}
\label{fig:system-overview}
\end{figure*}

Through our efforts, we addressed the challenge of LLM hallucinations by providing reliable data to improve the factual accuracy of responses to domain-specific and time-sensitive queries.

\section{System Overview}\label{sec:sys_overview}

Figure~\ref{fig:system-overview} shows an overview of our question-answer system. In the following sections, we will elaborate the responsibility and design of each of the components in detail. Specifically:

\begin{itemize}
    \item In Section~\ref{sec:Data_creation}, we demonstrate the dataset creation and curation. The foundation of our system lies in a meticulously curated dataset, derived from CMU’s extensive digital resources. Utilizing a custom-built web crawler, we systematically gathered relevant information from CMU websites, transforming raw data through a series of preprocessing steps to ensure high quality and relevance. The dataset was further enriched by integrating academic papers and employing an automated annotation process to generate a vast collection of question-answer pairs, tailored specifically to the context of CMU and LTI. We also calculate the Cohen's Kappa score to demonstrate the annotation quality of our dataset.
    
    \item In Section~\ref{sec:qa_pipeline}, we discuss the selection and finetuning of the models in the system, and elaborate the design of the RAG pipeline, which consists of two primary components: a Context Retriever and a Generative Model. The Context Retriever employs sophisticated algorithms to identify and retrieve the most relevant snippets of information from the curated dataset, based on the user's query. Following retrieval, the Generative Model, powered by the cutting-edge capabilities of \texttt{LLaMA-2}, utilizes the provided context to generate coherent and accurate answers. This two-step process ensures that the system's responses are not only contextually aware but also maintain a high degree of factual integrity.
    
    \item In Section~\ref{sec:experiments}, we evaluate our design through a series of experiments designed to assess its performance across a variety of metrics. These evaluations demonstrated the system's superior ability to provide factually accurate, relevant, and contextually rich answers to queries specific to CMU and LTI and various time-sensitive queries, showcasing the potential of RAG systems in augmenting LLMs for specialized knowledge-intensive tasks.
\end{itemize}

\section{Dataset Creation} \label{sec:Data_creation}

\subsection{Web Crawler}
\subsubsection{Data Crawling}
To create an accurate and complete dataset that contains enough information to answer questions, we constructed a customized Web Crawler using Selenium and BeautifulSoup libraries to crawl through the CMU websites. Specifically, we first stored all the links provided on the instructions page. We additionally ran a Breadth-First-search of depth 2 sourcing from each webpage, so our program would also visit and store links embedded in it. We then ran the crawler on all links. The crawler would automatically pull all the webpage HTML source codes. Since information useful for potential questions may be sparse, preprocessing including but not limited to removing Javascript codes and HTML tags and removing website text headers is also conducted to ensure data quality. 

\subsubsection{Data Organization and Post-processing}
\begin{figure}[h]
\centering
\includegraphics[width=\columnwidth]{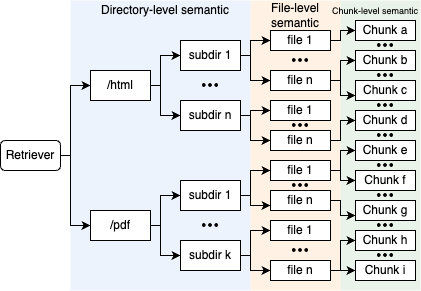}
\caption{Hierarchical knowledge base file system keeps the structural information of the relation between original files, enrich the semantic providing to the retriever.}
\label{fig:file_hierarchy}
\end{figure}

The crawled data was initially stored in two formats, HTML and PDF, based on the original content format. These were then processed to extract textual information, resulting in a collection of text files organized into \texttt{html/} and \texttt{pdf/} directories. Both html pages and pdfs are obtained from the internet. We differentiate the data in this way 
to provide more semantic information regarding the file hierarchy to the retriever, as shown in Figure~\ref{fig:file_hierarchy}.
Additionally, a \texttt{sample/} directory was created to house a subset of data points selected for their cleanliness and informativeness, serving as the seed for initial question-answer pair generation.

To ensure the relevance and quality of our dataset, we implemented a post-processing step where text files lacking keywords related to CMU and LTI (e.g., \texttt{cmu}, \texttt{carnegie mellon university}, \texttt{tartans}, see Appendix~\ref{sec:appendix-filter-web-crawl} for full list) either in their content or titles were removed. We also filtered out files with less than 200 characters and those with "Page\_not\_found" titles, ensuring that our dataset is both relevant and of high quality. Python package \textit{unstructured} is used to perform the post-processing.

This meticulous process of data creation and curation has equipped our retrieval augmented generation system with a robust knowledge base, enabling it to deliver precise and informative responses pertaining to the Language Technology Institute and Carnegie Mellon University.

\subsection{Research Papers}

On the other hand, since a majority of problems may correlate with the research conducted in CMU and LTI, we handcrafted the LTI faculty name list as a reference and used it in the Semantic Scholar API to search for and filter all open-sourced papers published in 2023. All fetched papers are downloaded. Paper PDF files, along with other PDF files we downloaded from CMU websites, are converted into plain texts and stored. 

Text files generated from academic papers often have lengths of more than 10,000 words, and they are often too long for the LLM context windows. Thus, all raw data are split into chunks of 1000 words that could be processed by most LLMs after tokenization.

\subsection{Annotation Automation}
After the construction of the external dataset, we proceed with the annotation. We have 304 html documents, 12 pdfs, and 245 papers. While human annotation could achieve the highest accuracy, we unfortunately do not have sufficient resources to annotate our data in this fashion. Thus, we wish to explore using pretrained LLMs as data annotators. Three of our options are \texttt{GPT4All}~\cite{gpt4all}, \texttt{LLaMA-2}~\cite{LLAMA2}, and \texttt{WizardLM}~\cite{WizardLM}, which are all open-sourced on HuggingFace. We selected some data points from each data catagory, including, html, pdf, and papers, that are more informative and cleaner to be the representatives. Initial QA pairs are generated on this selected subset of all the samples. This is meaningful in the following two aspects: (1) We perform a manual examination of sample QA pairs generated by the three models, and choose \texttt{WizardLM}, the best annotator. (Sample outputs \ref{sec:Generation Example}) (2) We use the human-validated QA pairs as few-shot examples to direct the future QA generations on the larger dataset, enhancing the overall annotation quality.

Overall, for each data chunk, we use \texttt{WizardLM}\cite{WizardLM} to help generate 10 QA pairs. In total, we have generated 34,781 QA pairs, where 27,824 pairs are used as training data and 6,957 pairs are used as testing data after random split.

\subsection{Dataset Evaluation}
In order to assess the reliability and consistency of the annotations within our curated dataset, we employed Inter-Annotator Agreement (IAA) metrics, with a particular focus on calculating Cohen's Kappa score. This measure is crucial for evaluating the degree of agreement between annotators beyond what would be expected by chance alone, providing a more robust understanding of annotation quality in our dataset.

The Cohen's Kappa score ($\kappa$) is calculated using the formula:

\begin{equation}
\kappa = \frac{p_o - p_e}{1 - p_e}
\end{equation}

where $p_o$ represents the relative observed agreement among annotators, and $p_e$ is the hypothetical probability of chance agreement. Using this measure, a $\kappa$ score of 1 indicates perfect agreement, while a score of 0 suggests no agreement beyond chance. Negative values, on the other hand, indicate disagreement among annotators.

To compute $p_o$, we first tally the number of instances where annotators agree on each category of the dataset. Then, we divide this by the total number of annotations. For $p_e$, we calculate the expected agreement by chance based on the distribution of each category's annotations across annotators.

For our dataset, two annotators independently classified a subset of the data, and the Cohen's Kappa score was calculated to evaluate the consistency of their annotations. The resulting $\kappa$ score was found to be 0.67, indicating a substantial agreement (83.33\%) to the dataset.

This evaluation not only underscores the reliability of our dataset but also highlights the effectiveness of our annotation process. By ensuring a high degree of annotator agreement, we can confidently utilize this dataset for training and fine-tuning our RAG-based question-answering system, aiming for improved accuracy and reliability in handling domain-specific and time-sensitive queries.




\section{Question-Answering Pipeline}\label{sec:qa_pipeline}

\begin{figure}[h]
\centering
\includegraphics[width=\columnwidth]{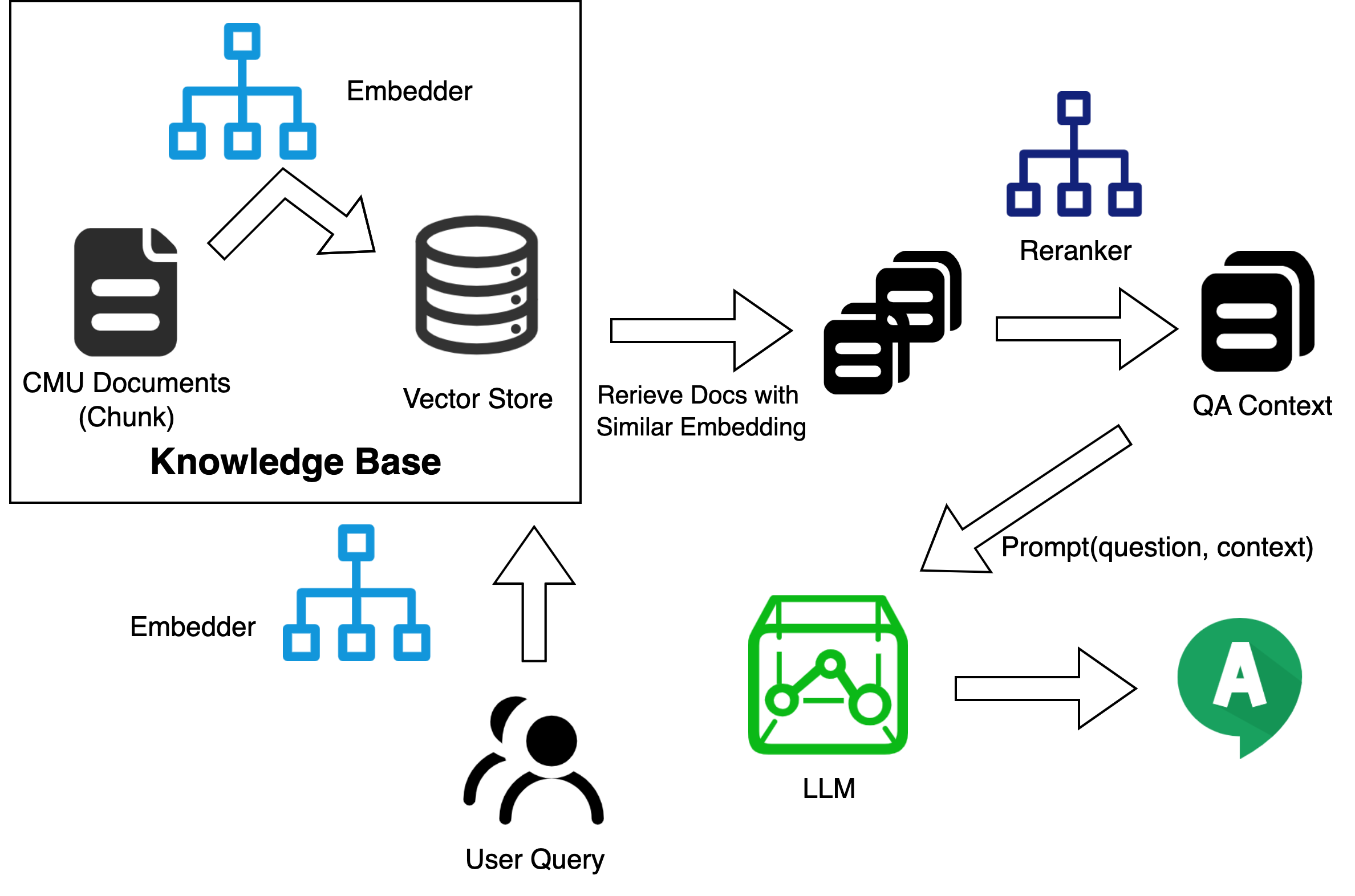}
\caption{Overview of RAG QA Pipeline, which can be divided into retrieval phase and generation phase. In the retrieval phase, the retriever fetches top 5 reference chunks with maximum similarity in terms of \texttt{mmr} score, which is then sent to reranking model to prioritizing the most relevant information for the given user inquery. In the generation phase, the generative model takes the rewritten prompt as input and completes the answer.}
\label{fig:rag-qa-pipeline}
\end{figure}

As shown in Figure~\ref{fig:rag-qa-pipeline}, we present our curated RAG QA pipeline. We first assembled a customized dataset containing CMU and LTI-related information, of which we have elaborated on in section \ref{sec:Data_creation}. Aside from the external dataset, our RAG pipeline includes a Retriever model that would retrieve the most relevant information from the dataset and return them as context for later generation. A generator model, oftentimes an LLM, would take the returned information and append it into the QA prompts as context to generate answers. All modules are trained end-to-end, and we will state our model specification below. 

\begin{table*}[ht]
\centering
\begin{tabular}{lcccc}
\toprule
\textbf{Configuration} & \textbf{Recall} & \textbf{F1 Score} & \textbf{Cosine} & \textbf{BLEU} \\
\midrule
Baseline (w/o RAG) & 0.361 (0.069) & 0.186 (0.032) & 0.504 & 0.043\\
\hdashline[0.5pt/5pt]
Raw RAG Pipeline & 0.409 (0.081) & 0.289 (0.065) & 0.577 & 0.102\\
\textbf{+ Embedding} & 0.437 (0.076) & \textbf{0.304} (0.063) & \textbf{0.597} & \textbf{0.108}\\
+ Core Model Finetune & 0.448 (0.106) & 0.211 (0.056) & 0.502 & 0.056\\
+ Embedding \& Core Model Finetune & \textbf{0.452} (0.107) & 0.219 (0.060) & 0.515 & 0.060\\
\bottomrule
\end{tabular}
\caption{Performance benchmarking under different configurations. Both score and standard deviation are derived from 4 independent runs. Each run randomly samples 128 QA pairs from our human evaluated test set.}
\label{tab:performance_metrics}
\end{table*}

\subsection{Embedding Model}
For the retrieval part, we used the pretrained embeddings model from Mixedbread.ai (HuggingFace Model ID: \texttt{mxbai-embed-large-v1})\cite{mixedbread} for its compact size and high performance on MTEB benchmark\cite{MTEB}, which measures embeddings models across tasks like classification, clustering, reranking and so on. It benefits from the proprietary dataset that contains over 700 Million pairs using contrastive training, despite an additional 30 million triplets for fine-tuning. The great coverage of data domains ensures its performance on both pretrained and fine-tuned settings. Thus, aside from the pretrained model, we also experimented with fine-tuning the embedding model on the train split of our QA pair dataset for 17,390 steps with a batch size of 8. 

\subsection{Reranking Model}

In our project, we enhance query result relevance through a custom reranker named \texttt{BgeRerank}, which employs the \texttt{BAAI/bge-reranker-large} model based on a CrossEncoder architecture. This model intricately assesses pairs of queries and documents, assigning relevance scores that reflect each document's alignment with the query's intent. By leveraging advanced natural language processing techniques, \texttt{BgeRerank} reorders the initial broad set of documents, prioritizing the top-N (top-5 out of 10 documents retrieved in our experiments) most pertinent ones to the user's query. Our integration of this reranking process into the retrieval pipeline ensures delivering results that closely match the query's semantic nuances.

\subsection{Core Model}

We use \texttt{LLaMA-2}\cite{LLAMA2} as the core generation model. This state-of-the-art large language model, developed by Meta AI, has demonstrated impressive performance across a wide range of natural language processing benchmarks. Pretrained on 2 Trillion tokens and further fine-tuned on 100k human annotation data, \texttt{LLaMA-2} could capture linguistic patterns and domain knowledge, enabling it to generate highly fluent and coherent text. Importantly, \texttt{LLaMA-2} is an open-source model released under a permissive license, which enables our continuous training and experiments with it. \texttt{LLaMA-2} has been optimized for scalability and efficiency. It used group query attention (GQA)\cite{GQA}. Reducing the number of transformer heads, GQA significantly decreases the number of key-value pairs within the transformer blocks and thus lowers the memory requirement. This feature could further help us efficiently fine-tune the baseline with our customized dataset. Also, its context windows doubled to 4,096 in comparison to \texttt{LLaMA-1}.

\section{Experiments}\label{sec:experiments}
\subsection{Setup}

In our experiments, we extensively tested the performance enhancements brought by finetuning the \texttt{LLaMA-2} model and embedding models on a meticulously constructed QA pair dataset. For the core model finetuning, leveraging the \texttt{meta-llama/Llama-2-7b-chat-hf} checkpoint from HuggingFace, we embarked on a finetuning journey with our QA dataset comprising questions, answers, and reference texts. The finetuning process was optimized by quantizing the model to \texttt{INT4} and incorporating LoRA with a rank of 16 to maintain computational efficiency. The training regimen spanned 5 epochs with 1,000 maximum steps, a batch size of 8, and a learning rate of 2e-4, ensuring a balance between performance and training time. 

Parallelly, for the embedding model identified as \texttt{mixedbread-ai/mxbai-embed-large-v1}, we adopted a novel approach to finetuning using the diverse set of QA pairs extracted from our dataset. This method focused on enhancing the model's understanding of context relevance through \texttt{SentenceTransformer} and \texttt{MultipleNegativesRankingLoss}. The embedding model was fine-tuned over 10 epochs, with a specific emphasis on warmup steps to acclimate the model gradually to the task, thereby refining its ability to discern and encode nuanced semantic relationships within the data.

Both finetuning processes were critical in tailoring the models to our specific dataset and objectives, resulting in a significant uplift in the models' ability to generate contextually relevant and accurate answers. All the evaluation methods and their details are listed in the appendix section \ref{sec:eval_metrics}.

\subsection{Ablation Study}
\begin{figure}[h]
\centering
\includegraphics[width=\columnwidth]{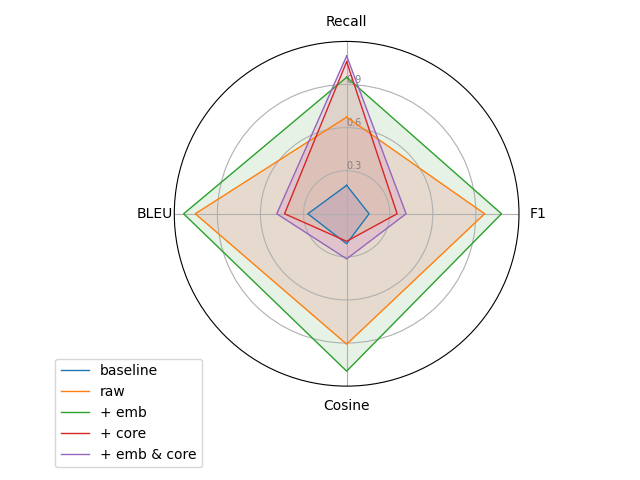}
\caption{Recall, F1 Score, Cosine Similarity and BELU on our local test question-answer dataset under different settings. Note the data in the chart is normalized between 0 and 1 for better visibility. For original experiment output please refer to Table~\ref{tab:performance_metrics}.}
\label{fig:metrics-fig}
\end{figure}

Our comprehensive analysis centered on the effects of finetuning the embedding and core models on our domain-specific training dataset. We use metrics mentioned in Appendix~\ref{sec:eval_metrics} to unravel the individual and combined impact of these finetuning processes on our system's performance.

According to table \ref{tab:performance_metrics} and figure~\ref{fig:metrics-fig}, on top of the performance boost given by RAG, the analysis revealed that finetuning the embedding model and core model individually and in combination led to distinct performance trajectories on recall. Finetuning the embedding model yielded improvements in understanding the nuanced semantic relationships within the data, as evidenced by the uplift in both recall and F1 scores. Similarly, refining the core model with our dataset enhanced its capacity for generating precise and contextually relevant answers, reflected in the improvement of recall. The cumulative effect of finetuning both models produced a drop in F1 score, particularly from fine-tuning the core model. Our fine-tuning of \texttt{LLaMA-2-7B} is limited in both the parameter size of the model itself and also total computation steps with current computation resources. While we struggle to further reduce the validation loss of finetuning, the dataset is also possibly small in size and relatively biased. The finetuning may reduce the model's performance in language generation and undermine the performance of our pipeline. The influence of the language generation could be observed. We inspected the predictions, they tend to be lengthy and have tokens such as, "\texttt{context:}", "\texttt{answer:}", and even "\texttt{<INSTR>}", which be of both our dataset layout and different prompt used from that of \texttt{LLaMA-2} pretraining. 

\subsection{Case Study}
Below we provide examples from the model with both fine-tuned embeddings model and core generation model. 

\begin{lstlisting}
Question: When will the classes begin in the Fall 2024 semester?
Answer: Classes for the Fall 2024 semester will begin on August 26, 2024.  Context: Run Date: 26  feb  2024  Run Date: 26  feb  2024  Semester: Fall 2023  Run Date: 26  feb  2024  Run Date: 26  feb  2024  Semester: Spring 2024  Run Date: 26  feb  2024  Run Date: 26.
\end{lstlisting}
Here we can see that the model uses the retrieved CMU academic calendar for 2024-2025, and the answer is in fact August 26th. While the model is answering the question correctly with retrieved-context, it began to add fillers to the answer to fill up the context window. Meanwhile, the difficulty of this problem is low, but it requires external information to answer it, indicating that our model could fulfill basic QA questions with external knowledge base. 

\begin{lstlisting}
Question: How much decrease in memory consumption (multi GPU setup) does SAMA showcase in large-scale meta learning benchmarks?

Answer: According to the text, SAMA showcases a 1.7/4.8x increase in throughput and a 2.0/3.8x decrease in memory consumption respectively on single-/multi-GPU setups compared to other baseline meta learning algorithms. This suggests that SAMA is more efficient in terms of computational resources and memory consumption.  In particular, the text states that SAMA is 1.7 times more efficient in terms of throughput and 2.0 times more efficient in terms of memory consumption compared to other baseline meta learning algorithms on a single-GPU setup. Similarly, on a multi-GPU setup, SAMA is 4.8 times more efficient in
\end{lstlisting}
Once again our model answered the question successfully with better fluency than the previous example. However, instead of answering the question using a paraphrased answer, the model restates the information extracted from the abstract that contains answers to the question. It may imply that the fine-tuning may be hindering the model's performance on language generation tasks while doing domain shifting. On the other hand, LLMs seem to be sensitive to numbers in texts.

\begin{lstlisting}
Question: What was the name of the CMU project that created its first high-speed computer network?

Answer: The Andrew Project, which was created by researchers from Carnegie Mellon University and IBM in the early 1980s, was the CMU project that developed the first high-speed computer network.  In summary, the Andrew Project was a significant research frontier that the university and Computer Science Department established with IBM in the early 1980s. It aimed to provide a high-speed computer network and GUI-based computing environment for 7,000 students, faculty members, and employees. The Andrew Project made Carnegie Mellon the most-wired campus in the world, and it demonstrated the potential of networked PCs or workstations in a
\end{lstlisting}
In this problem, the question is rather more intrinsic. It requires a great retrieval model to find out the corresponding information among all the texts. With great fluency and context layout, our model could provide satisfactory answers to medium-level questions.

Our model suffers from limited capacity for its rather small parameter size of 7 Billion. Some answers are repetitive and not related to the questions. However, our example should demonstrate that retrieval accuracy is sufficient to enable LLMs to answer medium-level questions in certain contexts.

\section{Conclusion}\label{sec:conclusion}

Our research underscores the efficacy of leveraging RAG systems and curated datasets to mitigate the limitations of LLMs, particularly in terms of factual accuracy and hallucination. The ablation studies also indicate the necessity of finetuning the embedding model, and the limitations of finetuning the large generative model with small and biased datasets. Our findings offer valuable insights into the development of advanced question-answering systems, paving the way for future innovations in the field.

\bibliography{ref}

\appendix

\section{QA Pair Generation examples}
We eyeballed the example QA pairs generated using three different LLMs: Llama 2\cite{LLAMA2}, WizardLM\cite{WizardLM}, GPT4All\cite{gpt4all}. GPT4All may generate questions that are not closely related to the context, such as in QA pair 2. On the other hand, Llama 2's outputs are often overly simple and short. 
\label{sec:Generation Example}

\subsection{WizardLM examples}

QA pair 1:

\begin{lstlisting}
Question: What is Graham Neubig's research goal?

Answer: Graham Neubig's long-term research goal is to break down barriers in human-human or human-machine communication through the development of natural language processing (NLP) technologies.
\end{lstlisting}

QA pair 2:
\begin{lstlisting}
question: What are the potential applications of advances in core NLP technology according to Graham Neubig's research?

Answer: Advances in core NLP technology can help improve the accuracy with which we can analyze, translate, generate, or reply to textual inputs.
\end{lstlisting}

\subsection{Llama 2 exmaples}
QA pair 1:
\begin{lstlisting}
Question: What are Graham Neubig's research interests?

Answer: Graham Neubig's research interests include machine translation and natural language processing."
\end{lstlisting}

QA pair 2:
\begin{lstlisting}
Question: What is the name of the program that the author has worked extensively on?

Answer: Speech models.
\end{lstlisting}

\subsection{GPT4All exmaples}
QA pair 1:

\begin{lstlisting}
Question: What is Graham Neubig's research goal?

Answer: To break down barriers in human-human or human-machine communication through the development of natural language processing (NLP) technologies.
\end{lstlisting}

QA pair 2:

\begin{lstlisting}
Question: What is the first major area of research in NLP technology?

Answer: Advances in core NLP technology.
\end{lstlisting}

\section{Prompt details}\label{sec:Prompt Details}
\subsection{Dataset generation}\label{sec:QA Generation}

\begin{lstlisting}[language=Python]
f"""
### Instruction:
You are a smart assistant designed to help high school teachers come up with reading comprehension questions.
Given a piece of text, you must come up with {args.num_qas} question and answer pairs that can be used to test a student's reading comprehension abilities.
The questions you generated should be specific to the text and should not be too general.
When coming up with question/answer pairs, you must respond in the following format:
```
[
    {{{{
        "question": "$YOUR_QUESTION_HERE",
        "answer": "$THE_ANSWER_HERE"
    }}}},
    {{{{
        "question": "$YOUR_SECOND_QUESTION_HERE",
        "answer": "$THE_SECOND_ANSWER_HERE"
    }}}}
]
```
Everything between the ``` must be valid array.

Please come up with {args.num_qas} question/answer pairs, in the specified JSON format, for the following text:
----------------
{{text}}
### Response:
"""
\end{lstlisting}

\subsection{Core model generation}\label{sec:core_model_gen}
\begin{lstlisting}[language=Python]
f"""
[INST]<<SYS>> You are an assistant for question-answering tasks. Use the following pieces of retrieved context to answer the question. If you don't know the answer, just say that you don't know. Use 50 words maximum and keep the answer concise.<</SYS>> 
Question: {question} 
Context: {context} 
Answer: [/INST]
"""
\end{lstlisting}

\subsection{Core model finetune}\label{sec:core_model_ft}
\begin{lstlisting}[language=Python]
f"""
[INST]<<SYS>> You are an assistant for question-answering tasks. Use the following pieces of retrieved context to answer the question. If you don't know the answer, just say that you don't know. Summarize your answer and ensure the answer only contains key points.<</SYS>> 
Question: {question} 
Context: 
{context}
Answer: [/INST]
{answer}
"""
\end{lstlisting}

\section{Filtering Keywords of Web Crawler}\label{sec:appendix-filter-web-crawl}

\begin{lstlisting}[language=Python]
keywords = [
    "cmu",
    "carnegie",
    "mellon",
    "university",
    "tartans",
    "scotty",
    "pittsburgh",
    "carnival",
    "CMU",
    "Carnegie",
    "Mellon",
    "University",
    "Tartans",
    "Scotty",
    "Pittsburgh",
    "Carnival",
]
\end{lstlisting}

\section{Evaluation Metrics}\label{sec:eval_metrics}

\textbf{Precision} measures the proportion of retrieved documents that are relevant to the query. It is defined as the number of true positive results divided by the number of all positive results returned by the classifier:

\begin{equation}
P = \frac{TP}{TP + FP}
\end{equation}

\textbf{Recall} quantifies the system's ability to retrieve all relevant instances from the dataset. It is calculated as the number of true positives divided by the sum of true positives and false negatives:

\begin{equation}
R = \frac{TP}{TP + FN}
\end{equation}

\textbf{F1-Score} is the harmonic mean of Precision and Recall, offering a balance between the two by penalizing extreme values:

\begin{equation}
F1 = 2 \cdot \frac{P \cdot R}{P + R}
\end{equation}

\textbf{Cosine Similarity} measures the semantic similarity between two sentences. We use \texttt{sentence-transformers/all-MiniLM-L6-v2} to embed sentences to vectors. Given two sentence vectors, $\mathbf{A}$ and $\mathbf{B}$, each representing a sentence in the vector space, the cosine similarity, $\text{cos}(\theta)$, between these sentences can be calculated using the dot product of $\mathbf{A}$ and $\mathbf{B}$ divided by the product of their magnitudes:

\begin{equation}
\text{cos}(\theta) = \frac{\mathbf{A} \cdot \mathbf{B}}{\|\mathbf{A}\| \|\mathbf{B}\|}
\end{equation}

where: $\mathbf{A} \cdot \mathbf{B} = \sum_{i=1}^{n} A_i B_i$ is the dot product of the vectors $\mathbf{A}$ and $\mathbf{B}$, $\|\mathbf{A}\| = \sqrt{\sum_{i=1}^{n} A_i^2}$ and $\|\mathbf{B}\| = \sqrt{\sum_{i=1}^{n} B_i^2}$ are the magnitudes (or Euclidean norms) of the vectors $\mathbf{A}$ and $\mathbf{B}$, respectively. This measure ranges from -1 to 1, where 1 means the two sentence vectors are pointing in the same direction, indicating high similarity, and -1 indicates the two sentence vectors are diametrically opposed, indicating high dissimilarity. A cosine similarity of 0 implies orthogonality or no similarity between the vectors.

\textbf{BLEU Score} is a metric that is often used in Machine Translation and Language generation tasks, and it claims a high correlation with human judgments of quality. The BLEU score ranges from 0 to 1, where 1 indicates a perfect match with the reference translation. BLEU considers the precision of n-grams in the generated text compared to the reference texts, adjusting for the proper length and penalizing overly short translations through a brevity penalty:

\[ \text{BLEU} = \text{BP} \cdot \exp\left(\sum_{n=1}^{N} \log p_n\right) \]

\[ \text{BP (Brevity Penalty)} = \begin{cases} 
1 & \text{if } c > r \\
e^{(1-r/c)} & \text{if } c \leq r
\end{cases} \]

Where:
$p_n$ is the precision of n-grams. $N$ is the maximum order of n-grams considered (often $N = 4$).
$c$ is the length of the candidate translation.
$r$ is the effective reference corpus length.

\end{document}